\documentclass[letterpaper, 10 pt, conference]{ieeeconf}  
\pdfoutput=1

\usepackage{hyperref}
\usepackage{listings}
\usepackage{graphicx}
\usepackage{epstopdf}
\usepackage{rotating}
\usepackage{multirow}
\usepackage{todonotes}
\usepackage{subcaption}
\usepackage{amsmath}
\usepackage{flushend}
\IEEEoverridecommandlockouts                             
\newcommand{\code}[1]{\texttt{\small{#1}}}

\title{\LARGE \bf
Decentralized collaborative transport of fabrics using micro-UAVs
}

\author{Ryan~Cotsakis, David~St-Onge and Giovanni~Beltrame
\thanks{M. Cotsakis is an undergraduate student from the University of British Colombia,
2329 West Mall, Vancouver, BC.
Dr. St-Onge  and Prof.
Beltrame are with the Department
of Computer and Software Engineering, \'Ecole Polytechnique de
Montr\'eal, 2900 Boul \'Edouard-Montpetit, Qu\'ebec
CA. e-mail: (\emph{name.surname}@polymtl.ca).}}

\begin{document}

\maketitle
\thispagestyle{empty}
\pagestyle{empty}

\begin{abstract}
  Small unmanned aerial vehicles (UAVs) have generally little capacity to
  carry payloads. Through collaboration, the UAVs can increase their joint payload
  capacity and carry more significant loads. For maximum flexibility to
  dynamic and unstructured environments and task demands, we propose a fully
  decentralized control infrastructure based on a swarm-specific scripting
  language, Buzz. In this paper, we describe the control infrastructure and
  use it to compare two algorithms for collaborative transport: field
  potentials and spring-damper. We test the performance of our approach with a
  fleet of micro-UAVs, demonstrating the potential of decentralized control
  for collaborative transport.
\end{abstract}


\section{INTRODUCTION}
\begin{figure} \centering
\includegraphics[width=0.9\linewidth]{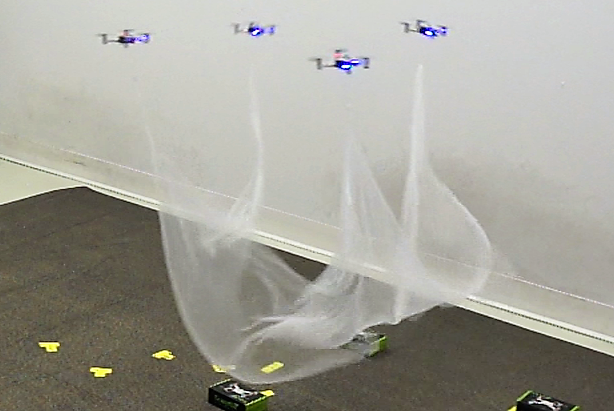}
\caption{Four Crazyflie micro-UAVs collaborating to transport a white fabric without requiring any central controller.}
\label{fig:4cf}
\end{figure}

Transport tasks for industrial, commercial and medical applications are now
considering small-scale aerial technologies ready for deployment. New use
cases are emerging, but the available payload remains the main limitation for
their implementation. Quite a novel use case was presented to our team from a
fashion designer: to involve micro-UAVs in the domestic task of getting
dressed. This proposal is challenging at many levels, namely: interaction,
clothing design, and UAV control. In this paper, we address the first piece of
the puzzle: how to use very small and therefore safe devices to transport
light pieces of clothing. Indeed, the proximity of the user and the narrow
space of a dressing room prevent the use of large devices. However, a decrease
in size tends to considerably decrease the payload. We propose to base our
solution on the use of multiple micro-UAVs collaborating as a whole for the
transport task, as shown in Fig.~\ref{fig:4cf}. The concepts required to
control the fleet are taken from swarm theory and its latest robotic
implementation, such as scalable behaviours based on local interactions.

The nature of the payload is itself a challenge for aerial transport. A flexible
material supported from many points creates a complex dynamic between the
carriers. Solving this problem has many other useful applications: 
supporting a net to catch other UAVs, transporting hoses to drop water over
fires, or slings for rescue missions on mountains have comparable dynamics
with bigger payload requirements. Nets, fabrics, hoses and slings, can all
be approximated with one or many Deformable Linear Objects (DLOs).

Since suspended fabrics are heavily influenced by the motion of air, the forces on the
UAVs are largely unpredictable, and can be quite large in magnitude.
In this case, the UAVs that share the payload must compensate for any
disturbances in real time. This can be achieved by having the UAVs cooperate,
and adjust their movement according to the positions and dynamics
(acceleration, torque, etc.) of their neighbours.


This paper is structured as follows: we discuss inspiring works in
Sec.~\ref{sec:rw}, then we detail our decentralized architecture in
Sec.~\ref{sec:dis} and we derive the control algorithms in
Sec.~\ref{sec:con}. Finally, the results of a set of experiments conducted with
Crazyflie micro-UAVs are presented and discussed in Sec.~\ref{sec:exp}.


\section{RELATED WORKS}
\label{sec:rw}
The transport of DLOs has been recently studied~\cite{Estevez2016}. The
feasibility of this task, with three UAVs, was demonstrated in simulation
using a particle swarm optimization to get the proper set of PID gains before
flight. This work, as with most of the literature of flexible payload
transport with UAVs, adopted a centralized controller. Similarly, the impressive
body of work of D'Andrea's group at ETH Zurich includes works such three UAVs
throwing and catching a ball in a net~\cite{Ritz2012} and the transport of a
flexible ring with six UAVs~\cite{Ritz2013}; but they rely on a fully
centralized and latency-free state estimator and controller.

\begin{figure*}[!h]
\centering
\begin{subfigure}[t]{0.48\linewidth}
\includegraphics[width=\linewidth]{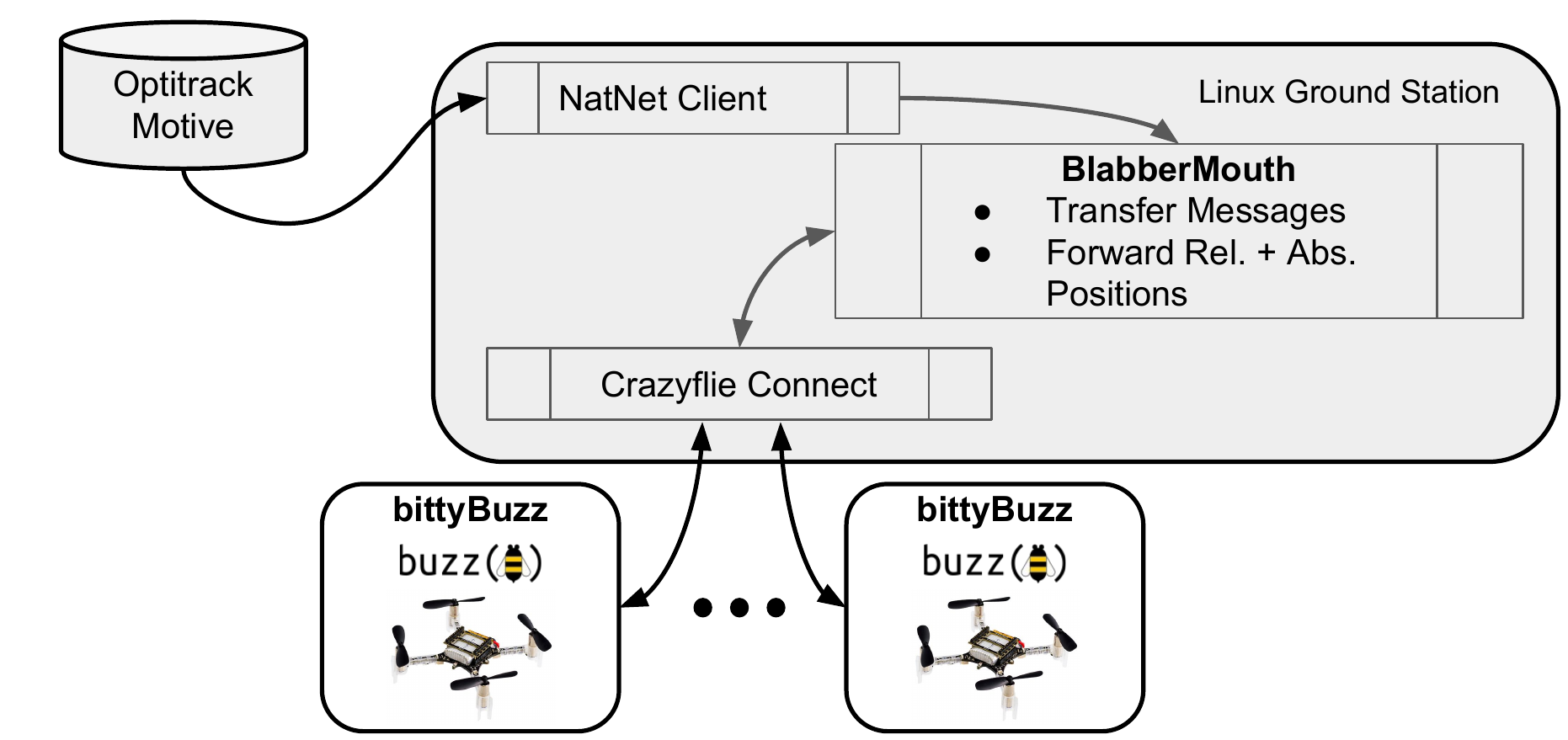}
\caption{}
\end{subfigure}
\begin{subfigure}[t]{0.48\linewidth}
\includegraphics[width=\linewidth]{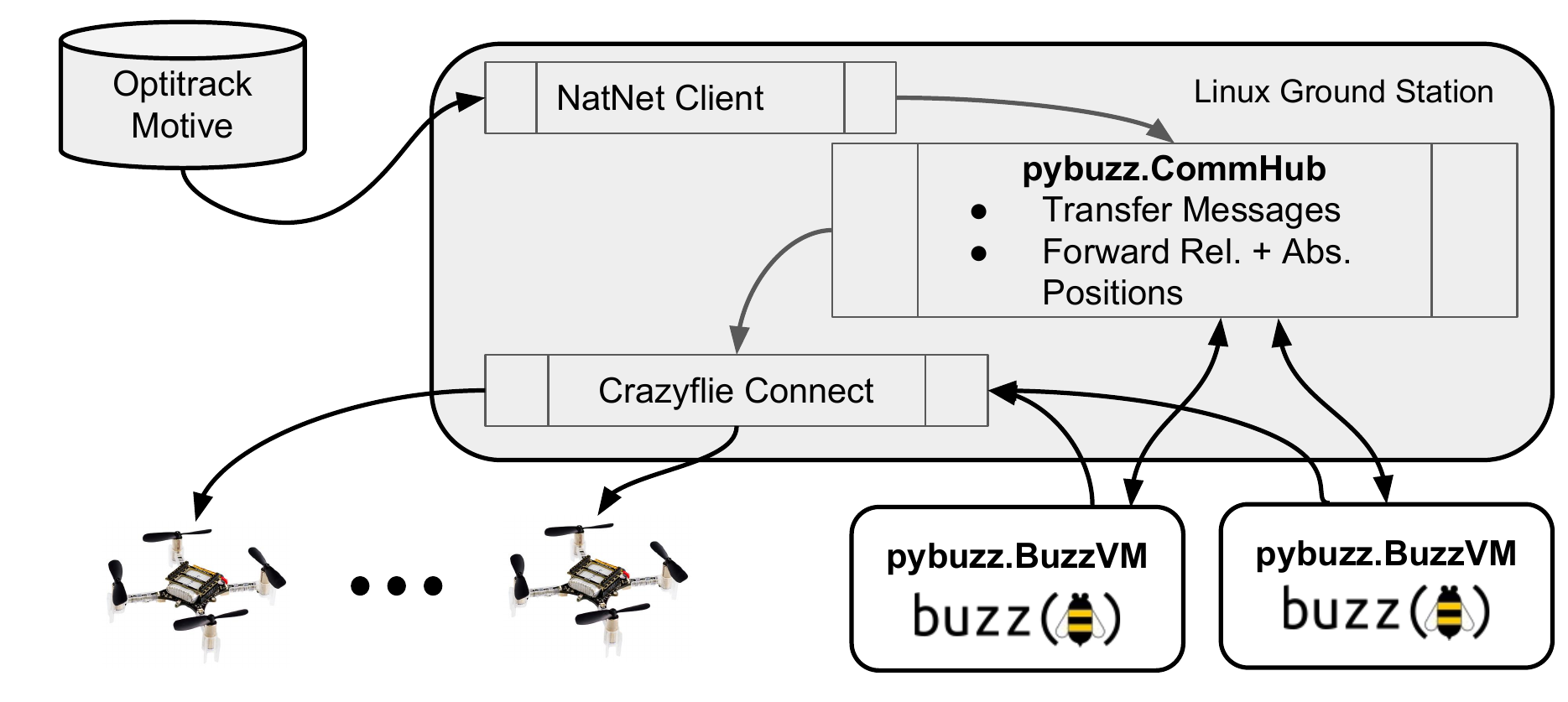}
\caption{}
\end{subfigure}
\caption{Buzz deployment architecture for the Crazyflie: \textbf{(a)} Buzz running on-board. \textbf{(b)} Buzz emulated using PyBuzz.}
\label{fig:arch}
\end{figure*}

Distributed or decentralized approaches to the problem are seen less
frequently. Distributed collaborative transport was achieved for wheeled robots,
pushing a rigid geometric object~\cite{Dai2016} with a combination of
potential field forces. More recently, two UAVs were designed to cooperate in
a fully decentralized configuration for the transport of a rigid body using
only inertial measurement and vision~\cite{Loianno2018}. Both are inspiring
works, but are still far from the target of this work.

The same type of micro-UAVs used in our experiments, the Crazyflie, was
subject to a detailed swarm controller design~\cite{Preiss2017}. We leverage
the stability of their on-board control and the scalability of their
configuration to deploy our solution. Nevertheless, collaborative payload
transport has never been attempted with such a small and sensitive device. Other
solutions for controlling a swarm of Crazyflies has been
considered~\cite{Furci2015,Honig2017}, but they are centralized. As with
Preiss et al., their scripts are specific to their implementation for this
platform, whereas we leveraged a domain-specific programming language for
swarms~\cite{Pinciroli2016b} to guarantee portability to other platforms.


\section{DECENTRALIZED ARCHITECTURE}
\label{sec:dis}

The aim of this project is not only to achieve a fleet of Crazyflies collaborating on
a transport task, but to study a behaviour that can the be ported to other
robotic platforms. The Buzz domain-specific language~\cite{Pinciroli2016b} and its
associated virtual machine was successfully deployed on 3DR Solos, DJI Matrice
100, Intel Aero, and Clearpath Husky through a ROS connector as well as on
K-Team Khepera, Kilobots and Zooids, natively.

%

\subsection{Buzz}
To accelerate the implementation of swarm behaviours, Buzz provides a set of
special primitives from which we leverage two essential concepts: a) swarm
aggregation and b) neighbour operations. We detail here these concepts to ease
the comprehension of the following implementation, but all information can be
found in previous publications specific to Buzz language~\cite{Pinciroli2016b}
and example scripts that are available
online\footnote{\url{http://the.swarming.buzz}}.

\textit{Swarm Aggregation} is a primitive which allows for grouping of robots
into sub--swarms, through the principle of dynamic
labelling~\cite{Pinciroli2016b}. The \code{swarm} construct is used to create a
group of robots which can be attributed with a specific behaviour that
differs from the other robots, based either on the task or robot abilities.

\textit{Neighbour Operations} in Buzz refer to a rich set of functions which
can be performed with or on neighbouring robots through situated
communication~\cite{Stoy2001}. Neighbours are defined from a
network perspective as robots which have a direct communication link
with each other. With situated communication, whenever a robot receives a
message, the origin position of the message is also known to the receiver.

Finally, any Buzz script is compiled into an optimized, memory-efficient, and
platform-agnostic bytecode to be executed on the Buzz Virtual Machine
(BVM). To interface the BVM with the robots' actuators and sensors, the
integrator needs to write his own C hooks that are callable from a Buzz script.

\subsection{PyBuzz}
Rather than embedding the BVM into the firmware of each robot,
we decided to accelerate the development of the control algorithms
by creating a centralized emulator for our decentralized configuration.
Emulating a distributed software architecture requires a wrapper that can be
instantiated to connect to each hardware node. To solve this problem, we
created a Python module for wrapping the BVM called PyBuzz, such that
in Python, one can construct a BVM as a Python object, and link Python
functions as callable functions in Buzz. When interpreting the Buzz object
code, the BVM performs calls to these Python functions.

By importing \code{pybuzz} in a Python script, one can construct any number of BVM objects.
The translation between C and Python was
achieved using Cython~\cite{Behnel2011}, a tool that is used for writing C
extensions for Python.

On a ground station, we have information about the locations of all of the
UAVs. These locations are fed into a communication hub, \code{pybuzz.CommHub},
as shown in Fig.~\ref{fig:arch}-b. The
communication hub emulates the robots' communication range, giving each BVM its
current location, and the relative locations of its neighbours. While each BVM
continuously steps through the Buzz script, the communication hub automatically
manages the transmission of all messages sent between each BVM. The only information sent to the Crazyflie's hardware is its current location (received from \code{pybuzz.CommHub}) and instructions for course adjustments (received from \code{pybuzz.BuzzVM}).

In the end, a Buzz script tested with the PyBuzz emulator (Fig.~\ref{fig:arch}-b)
can work unmodified on board a Crazyflie (Fig.~\ref{fig:arch}-a).


\section{DECENTRALIZED CONTROL}
\label{sec:con}
The accurate control of a fleet carrying a shared payload is a problem
usually approached from a robust low-level control
perspective~\cite{Ritz2013}. However, modelling the payload between the robots
as a set of local interactions can bring it up to the fleet coordination
level, for which decentralized solutions are widely studied in swarm behaviours.

\subsection{Flexible payload model}


We take inspiration from the modelling of hoses, approximated for UAVs collaborative payload with discrete linear objects~\cite{Estevez2015}. We also postulate that the transport is achieved only with UAVs at cruise heights, i.e. with the fabric freely hanging. However, our experiments show that this approximation can hold for stable take-off and landing while holding the fabric in certain conditions (see Sec.~\ref{sec:exp}).

\begin{figure}[!h] \centering
\includegraphics[width=0.7\linewidth]{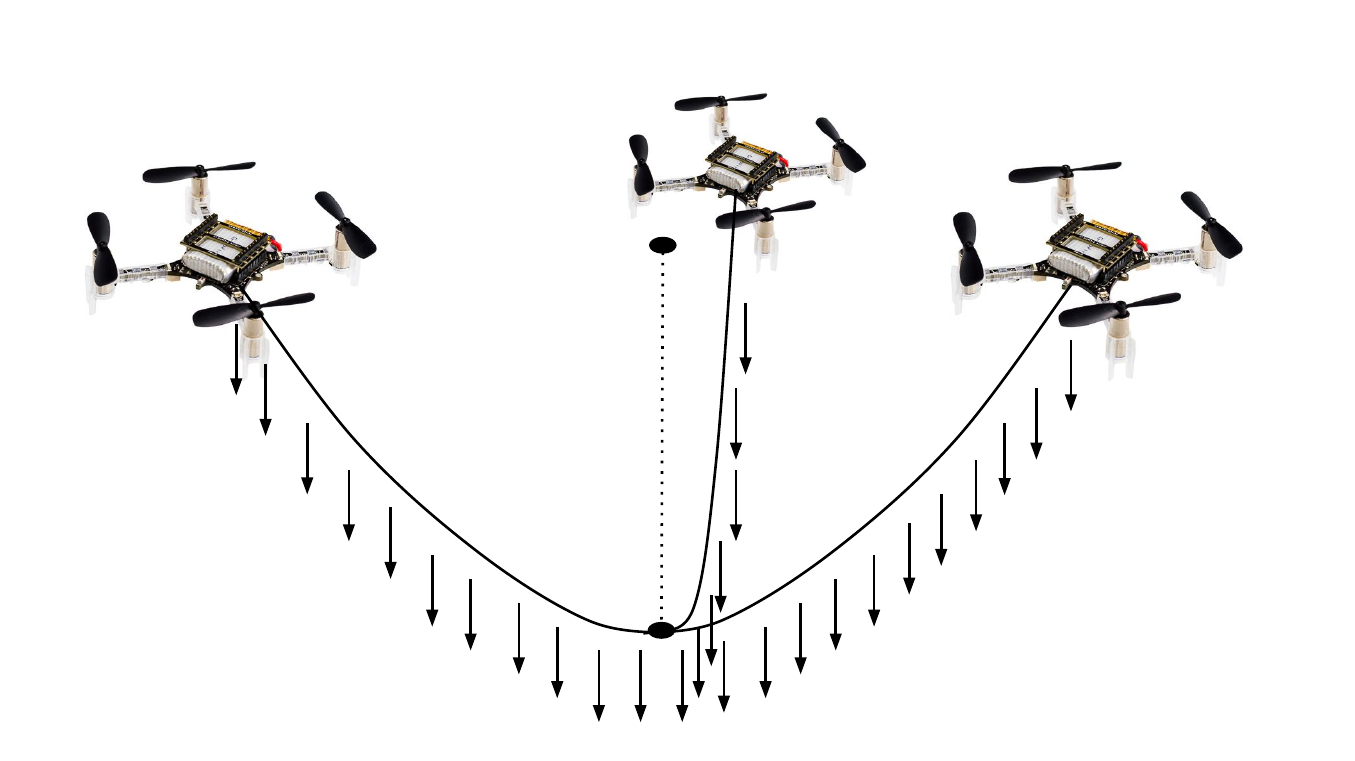}
\caption{Modelling fabric transport by connected DLOs. Each of the three Crazyflies carries half of a catenary curve with its vertex at a point directly below the centre of mass of the swarm}
\label{fig:DLO}
\end{figure}

The payload can be approximated by cables with uniformly distributed weight creating catenary curves that hang from each robot to a point directly below the centre of mass of the swarm, which we will refer to as the \emph{connecting point}. The vertex of each catenary curve coincides with the connecting point. This model is depicted in Fig.~\ref{fig:DLO}. With $n$ robots all at the same height sharing a payload of mass $m$, the vertical tensile force $T_z$ is approximately
\begin{equation}
T_z = \frac{mg}{n} 
\end{equation}
The horizontal force $T_x$ is a much more involved calculation. We recall the equation of a catenary curve to be
\begin{equation} \label{cat}
z(x) = a\cosh \frac{x}{a}
\end{equation}
where $a$ is a parameter that describes how taught the approximating cable is. Remaining under the assumption that the robots have the same altitude, $a$ can be solved for numerically in the transcendental equation
\begin{equation} \label{ax0}
L = a\sinh \frac{x_0}{a}
\end{equation}
where $L$ is the length of the approximating cable from the robot to the connecting point, and $x_0$ is the horizontal distance from the robot to the centre of mass of the swarm. If we take the derivative of Eqn.~\ref{cat}, we see that
\begin{equation}
\frac{T_z}{T_x} = z'(x_0) = \sinh \frac{x_0}{a} = \frac{L}{a}
\end{equation}
If we observe through Eqn.~\ref{ax0} that $a$ is a function of $x_0$, we can see that $T_x$ is a function of $x_0$ and other constants:
\begin{equation} \label{Tx}
T_x(x_0) = \frac{mg}{nL}a
\end{equation}

\subsection{Spring-Damper analogy}
\label{subsec:SD}
An intuitive method of having the UAVs maintain a stable formation is to simulate the forces due to springs and dampers emplaced between each robot. Spring and damper control was shown to work well in a simulated swarm~\cite{Belkacem2016}. The repulsive force between two robots as a function of the distance $d$ between them would be
\begin{equation} \label{SD}
F_{rep} = -k(d-l_0) - B\dot{d}
\end{equation}
where $k$ and $l_0$ are respectively the spring constant and unstretched length of the spring, and $B$ is the damping constant of the damper.

To tune the parameters $k$ and $B$, recall the horizontal force on each robot due to the payload from Eqn.~\ref{Tx}. Let us define $k_p$ to be the value of $T_x'$ at the equilibrium position of $x_0$. Then for small perturbations from the equilibrium position, the horizontal force due to the payload can be approximated by a spring with spring constant $k_p$ connecting the centre of mass of the swarm to the robot. By relating the payload to a spring force, we can interpret the entire system as a sum of forces from springs and dampers to simplify the analysis.

We can calculate the net force on a particular robot in the $x$ direction, the direction towards the centre of mass of the swarm, for a small change $\Delta x$ from the equilibrium position. We denote this force as $F_x$ and express it as a sum over the robot's $N$ neighbours:
\begin{equation} \label{Fx}
F_x = -\Big(k_p + k\sum_{i = 1}^{N}\cos^2\theta_i \Big)\Delta x
\end{equation}
where $\theta_i$ is the angle from the centre of mass of the swarm to neighbour $i$ with respect to the robot in question. In the same way we can calculate the force in the y direction resulting from a small perturbation $\Delta y$ to be
\begin{equation} \label{Fy}
F_y = -\Big(k\sum_{i = 1}^{N}\sin^2\theta_i \Big)\Delta y
\end{equation}
By substituting equations \ref{Fx} and \ref{Fy} with the values 
\begin{equation} \label{kxy}
k_x = k_p + k\sum_{i = 1}^{N}\cos^2\theta_i,\ \ k_y = k\sum_{i = 1}^{N}\sin^2\theta_i
\end{equation}
we are able to simplify the system of springs down to two springs. One parallel to the $x$-axis with spring constant $k_x$, and one parallel to the $y$-axis with spring constant $k_y$. What remains is to simplify the system of dampers.

Similarly to how we defined $k_x$ and $k_y$ in Eqn.~\ref{kxy}, we define
\begin{equation} \label{Bxy}
B_x = B\sum_{i = 1}^{N}\cos^2\theta_i,\ \ B_y = B\sum_{i = 1}^{N}\sin^2\theta_i
\end{equation}
and interpret our system separately in dimensions $x$ and $y$ when choosing parameters $B$ and $k$. In the $x$ direction we have, attached to the Crazyflie's mass, a spring and a damper in parallel with parameters $k_x$ and $B_x$ respectively. In the $y$ direction, we have a spring and damper in parallel with parameters $k_y$ and $B_y$ respectively. We tune $B$ and $k$ such that both of these systems in each dimension are nearly critically damped. 

The calculation of $\dot{d}$ in Eqn.~\ref{SD} is not a straight forward task when $d$ is being sampled discretely in time and space. To mitigate the errors induced by the lack of continuity, we use all $n$ previously measured values of $d$, and weigh each measurement's importance with a coefficient that's magnitude decays exponentially with time. The simple implementation of this algorithm is to initialize another distance variable $\tilde{d}$ that acts as a representative of all previous values of $d$.

Given that at time $t_n$ we measure the distance between robots to be $d_n$, we define
\begin{equation}\label{td}
\tilde{d}_n = w_n \tilde{d}_{n-1} + (1 - w_n) d_n, \text { with } \tilde{d}_0 = d_0
\end{equation}
and
\begin{equation}
w_n = \exp \left(-\frac{t_n - t_{n-1}}{\tau}\right)
\end{equation}
where $\tau$ is a time constant that governs how quickly information from previous measurements is lost. We aim to show that we can approximate the velocity as follows:
\begin{equation} \label{ddot:1}
\dot{d} = \frac{d_n - \tilde{d}_n}{\tau}
\end{equation}
In the case where $d$ is measured at regular time intervals $\Delta t$, then $w_n= w = \exp(-\Delta t / \tau),\ \forall n$ which gives a better approximation:
\begin{equation} \label{ddot:2}
\dot{d} = \frac{(1-w)(d_n - \tilde{d}_n)}{w\Delta t}
\end{equation}

To show that Eqn.~\ref{ddot:2} is a reasonable approximation, let us consider the explicit form of $\tilde{d}_n$:
\begin{equation} \label{dnsum}
\tilde{d}_n = (1 - w)\sum_{i=0}^\infty d_{n-i}w^i
\end{equation}
Since we aim to obtain an expression for $\dot{d}$, let us assume that it is relatively constant (at least for a short interval of time). This allows us to make the approximation
\begin{equation}
d_{n-i} \approx d_n - i\Delta t\dot{d}
\end{equation}
With this approximation, Eqn.~\ref{dnsum} can be expressed as
\begin{equation}
\tilde{d}_n = (1 - w)\sum_{i=0}^\infty (d_n - i\Delta t\dot{d})w^i
\end{equation}
which can be evaluated exactly:
\begin{equation}
\tilde{d}_n = d_n - \frac{w\Delta t\dot{d}}{1-w}
\end{equation}
Finally, solving for $\dot{d}$ yields Eqn.~\ref{ddot:2}.

With $w$ being a exponential function of $\frac{\Delta t}{\tau}$, it can easily be shown that as $\frac{\Delta t}{\tau} \rightarrow 0$,
\begin{equation}
\frac{1-w}{w\Delta t} = \frac{1}{\tau}
\end{equation}
which when substituting in Eqn.~\ref{ddot:2} implies Eqn.~\ref{ddot:1}. One downside to using Eqn.~\ref{ddot:1} is that we require $\tau \gg \Delta t$, and since $\dot{d}$ is an approximation of the velocity $\frac{\tau}{2}$ seconds in the past, our measurement of $\dot{d}$ is delayed. The lack of $w$ and $\Delta t$ dependence in Eqn.~\ref{ddot:1} implies that it can be used over Eqn.~\ref{ddot:2} whenever $d$ is sampled irregularly, as long as $\tau$ is sufficiently large.

\subsection{Bacteria interaction analogy}
\begin{figure} \centering
\includegraphics[width=\columnwidth]{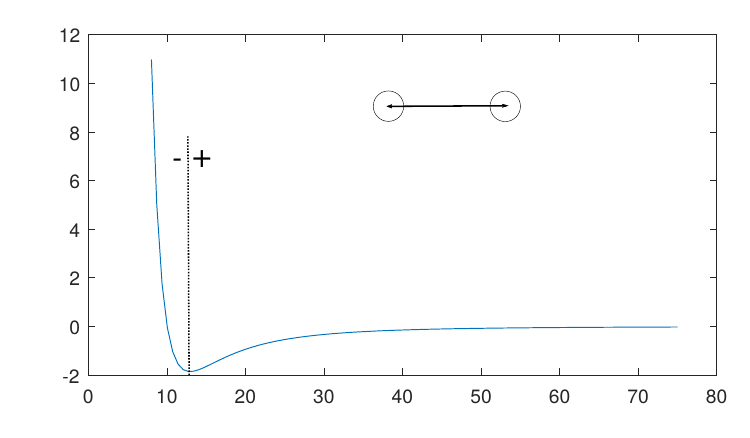}
\put(-150,88){\scriptsize
$LJ = \sum^n_{i=1}\frac{\epsilon \sigma^2}{8}[(\frac{\sigma}{d})^4-2(\frac{\sigma}{d})^2]$
}
\put(-130,2){\scriptsize
$d$ (cm)
}
\put(-105,110){\scriptsize
$d$} \put(-230,45){\begin{rotate}{90}\scriptsize Lennard-Jones
Potential\end{rotate}}
\caption{Lennard-Jones potential adapted
for Crazyflies. The '-' and '+' domains are
respectively the repulsive and attractive parts, for which the
pivot point is set with parameter $\sigma$. $d$ is the distance
between two robots and $\epsilon$ is a parameter acting as a
control gain on the potential.}
\label{fig:lj}
\end{figure}
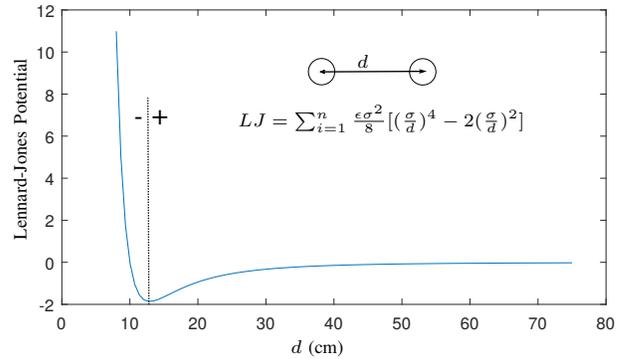

Among the most popular formalizations of biological swarm behaviours, potential
functions are a simple, yet flexible control approach.
Artificial potential functions have been used extensively for robot navigation
and control~\cite{Rimon1992,Reif1999}. Based on their knowledge of their neighbours' positions,
each robot computes a virtual force vector:
\begin{equation}
F=\sum^N_{i=1} f(d_i)e^{j\theta_i}
\end{equation}
where $\theta_i$ and $d_i$ are the direction and the distance to the $i$th
perceived obstacle or robot, and the function $f(d_i)$ is the negative gradient of an
artificial potential function. One of the most commonly used artificial potentials is
the Lennard-Jones potential, adapted for our physical system as shown in
Fig.~\ref{fig:lj}.

The two parts of the potential equation represent
the attractor and repulsor effect. This potential is driven by two parameters:
the target distance between robots $\sigma$ and the strength $\epsilon$ of the
potential.

Based on its popularity, we selected Lennard-Jones potential to control the behaviour of our robotic swarm:
\begin{equation} \label{lj}
LJ = \frac{\epsilon \sigma^2}{8}\left[\left(\frac{\sigma}{d}\right)^4-2\left(\frac{\sigma}{d}\right)^2\right]
\end{equation}
We can compute the resulting force exerted as the negative gradient of Eqn.~\ref{lj}:
\begin{equation}
F_{rep} = \frac{\epsilon \sigma}{2}\left[\left(\frac{\sigma}{d}\right)^5-\left(\frac{\sigma}{d}\right)^3\right]
\end{equation}
This representation of the Lennard-Jones repulsive force has a very elegant Taylor expansion about $d = \sigma$, that is
\begin{equation} \label{taylor}
F_{rep} = -\epsilon(d-\sigma) + O\left((d-\sigma)^2\right)
\end{equation}
The first term in Eqn.~\ref{taylor} has the same form as the first term in Eqn.~\ref{SD}. With $\epsilon = k$ and $\sigma = l_0$, the Lennard-Jones potential is, to second order, identical to a spring system at the equilibrium position and can be approximated by a harmonic oscillator. By representing the potential in this form, we are able to compare Lennard-Jones experiments with spring-damper experiments by using the same parameters derived in Sec.~\ref{subsec:SD}.


\subsection{Fleet translation}
With both control approaches, a goal (target location) is represented as an attractor influencing the whole
group. The final displacement vector at each step is then computed from a weighted sum of this attraction force and 
the forces resulting from the formation algorithm.
Under ideal circumstances, the robots move towards their final position at constant velocity.


\section{EXPERIMENTS}
\label{sec:exp}
To start, we devised an unorthodox method to have a swarm of Crazyflies be seen and distinguished by our motion sensing platform, Optitrack, since the UAVs are too small to allow for many unique configurations of reflective markers. Our solution was to attach a single marker on each Crazyflie, which provides the position of the UAV in 3-D space, but not the orientation. We are able to control the attitude of the Crazyflie with the on-board controller using its Inertial Measurement Unit. To have the robots be uniquely identifiable, we tell the software approximately where each Crazyflie is expected to start, and conduct a grid search before takeoff. For the duration of the flight, we assume continuous motion of the Crazyflie. This way the software can identify which robot is which based on the previous frame.

Building on the infrastructure work described in Sec.~\ref{sec:dis}, we implemented Buzz scripts for both algorithms. We conducted eleven flight tests, each involving three Crazyflies. The first of these tests demonstrates the flight of the three Crazyflies without any payload, nor any control algorithm facilitating the distances between robots. Five of the flight tests demonstrate the transport of fabric using the spring-damper algorithm, and the last five demonstrate the transport of fabric using the Lennard-Jones algorithm. All other variables were kept constant. In each flight test, the Crazyflies were instructed to take off from slightly elevated platforms to 80 cm and maintain this altitude while travelling horizontally at constant velocity. For the tests involving the transport of fabric, one of the corners of the fabric was manually tied to each Crazyflie with fishing line before takeoff.

\subsection{Weight limit}
From BitCraze specifications, a Crazyflie can lift 42 g and its own weight is 27 g, leaving 15 g for a payload, from which we already use 0.1 g for an Optitrack marker. In order to confirm and test these numbers, we measured the maximum lifting capability of the Crazyflies. Using a digital spring scale with a resolution of 1 g, we took 5 measurements of the maximum payload, and all measurements gave 18 g.

\subsection{Spring-damper control versus Lennard-Jones potential}

%

\begin{figure}[!h] \centering

\begin{subfigure}[t]{0.9\linewidth}
\includegraphics[width=\linewidth]{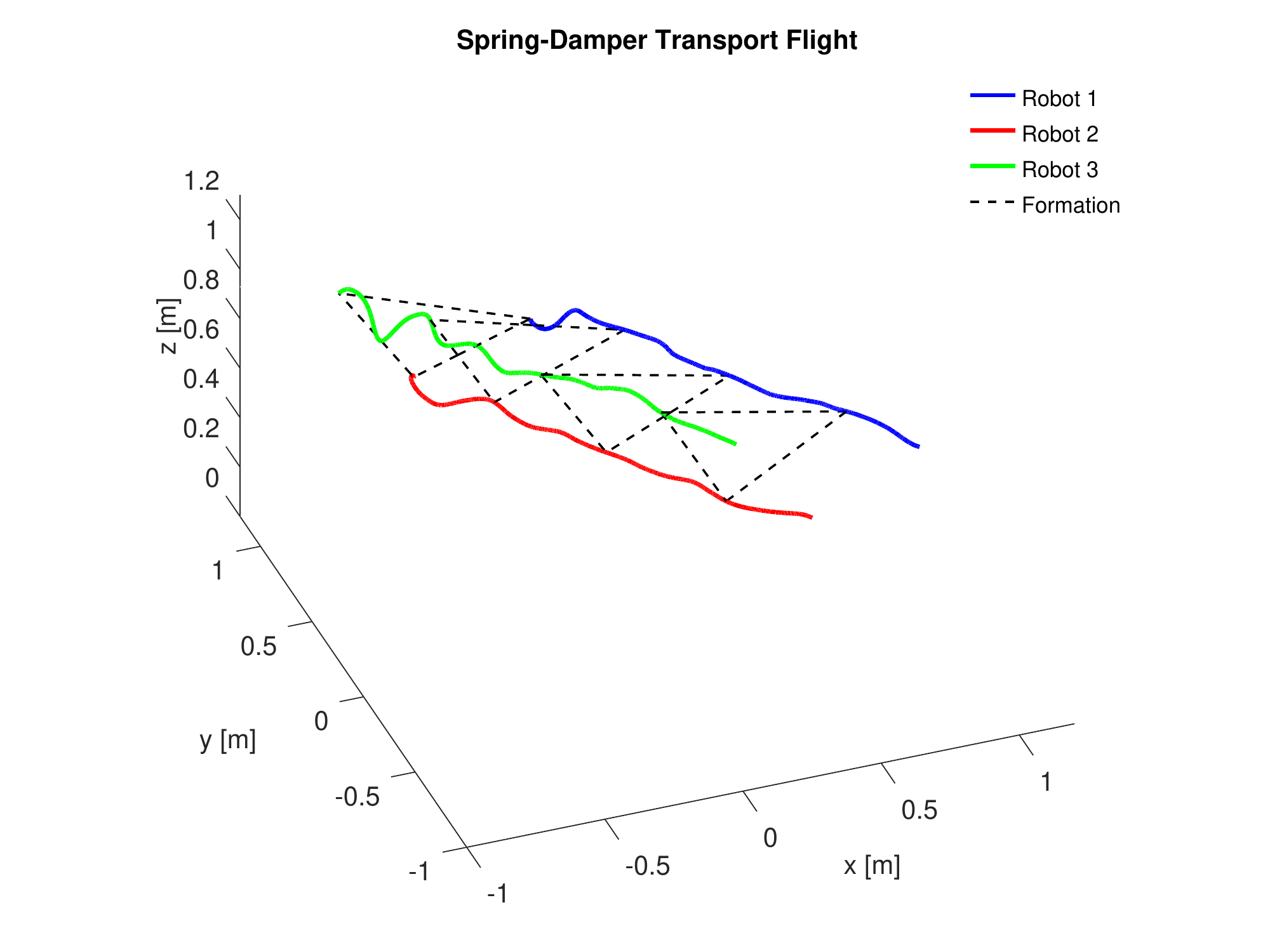}
\caption{}
\end{subfigure}

\begin{subfigure}[t]{0.9\linewidth}
\includegraphics[width=\linewidth]{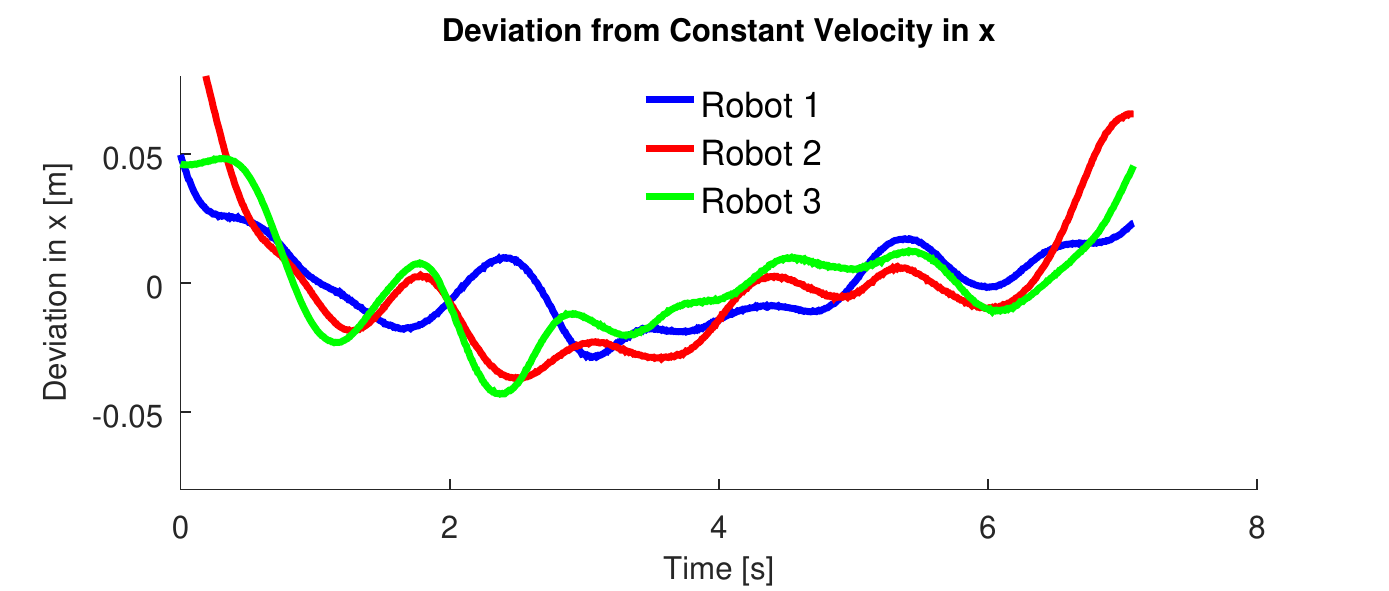}
\caption{}
\end{subfigure}
\begin{subfigure}[t]{0.9\linewidth}
\includegraphics[width=\linewidth]{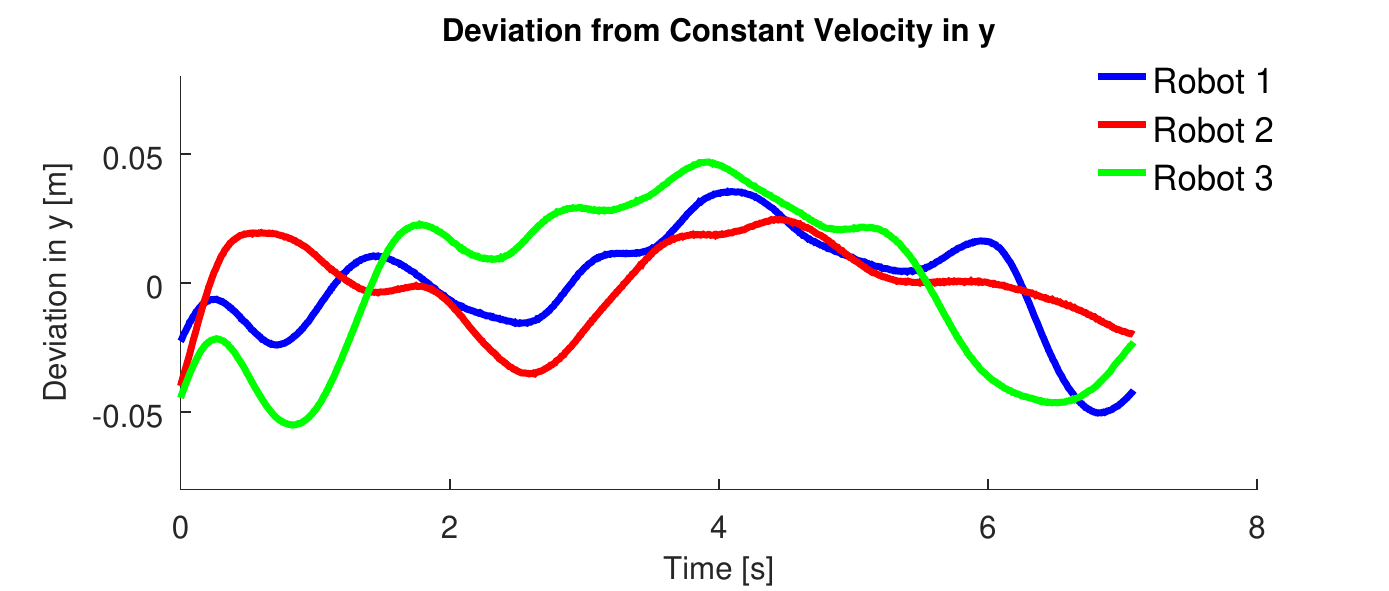}
\caption{}
\end{subfigure}
\caption{\textbf{Spring-damper} flight test: \textbf{(a)} Overview of flight path of the three Crazyflies while cooperatively carrying fabric during one of the 5 flight tests of its kind. The vertices of the dotted triangles indicate where the Crazyflies are at 2 second intervals. \textbf{(b)} Deviation of each robot's $x$ coordinate from the best approximating constant velocity trajectory for the curves in (a). Each curve is a plot of $x$ vs $t$ with the zeroth and first order terms subtracted, i.e. the average value and average slope of each curve is 0. \textbf{(c)} Exactly the same construction as (b), but taking the $y$ vs $t$ graph, rather than the $x$ vs $t$.}
\label{fig:SD3}
\vspace{-1.0em}
\end{figure}

\begin{figure}[!h] \centering

\begin{subfigure}[t]{0.9\linewidth}
\includegraphics[width=\linewidth]{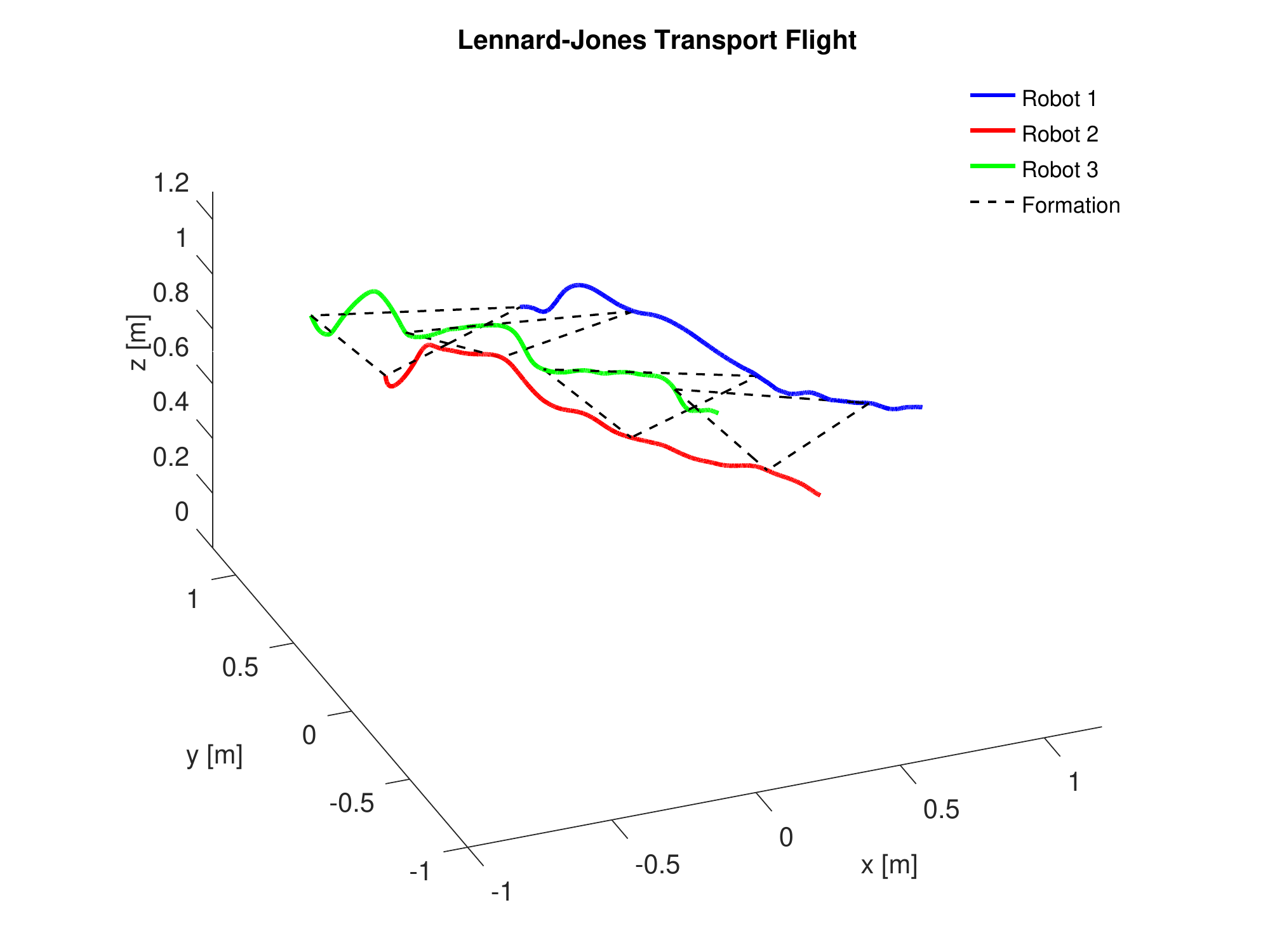}
\caption{}
\end{subfigure}\\

\begin{subfigure}[t]{0.9\linewidth}
\includegraphics[width=\linewidth]{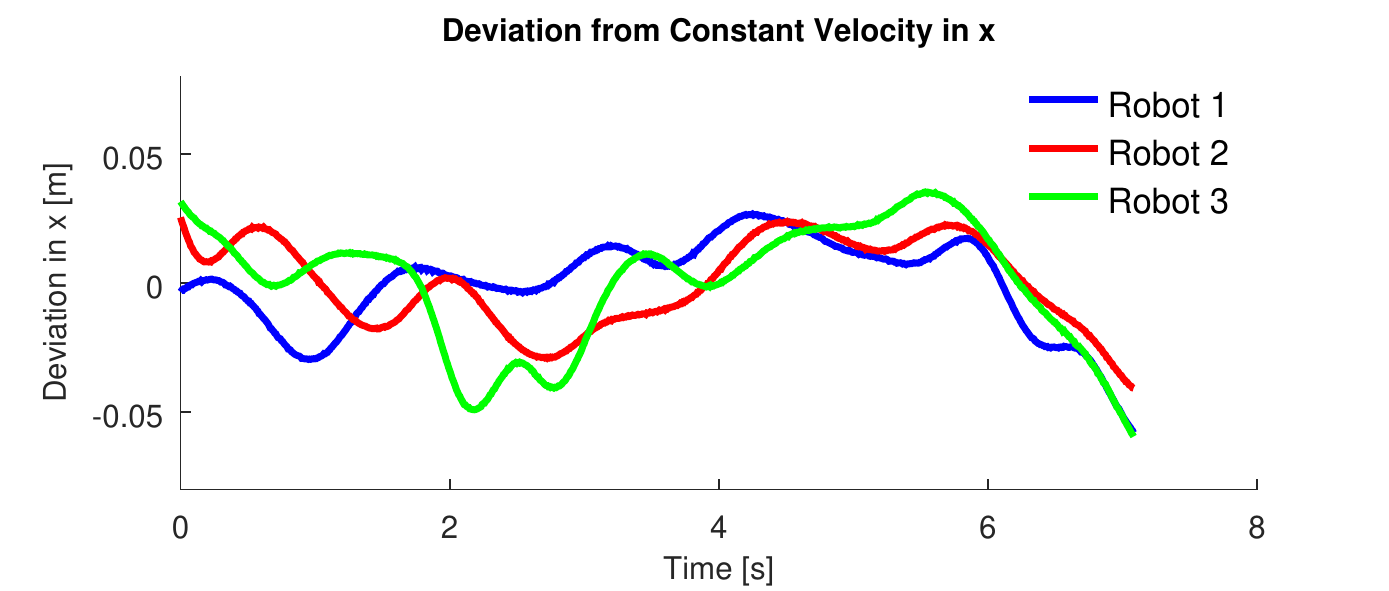}
\caption{}
\end{subfigure}
\begin{subfigure}[t]{0.9\linewidth}
\includegraphics[width=\linewidth]{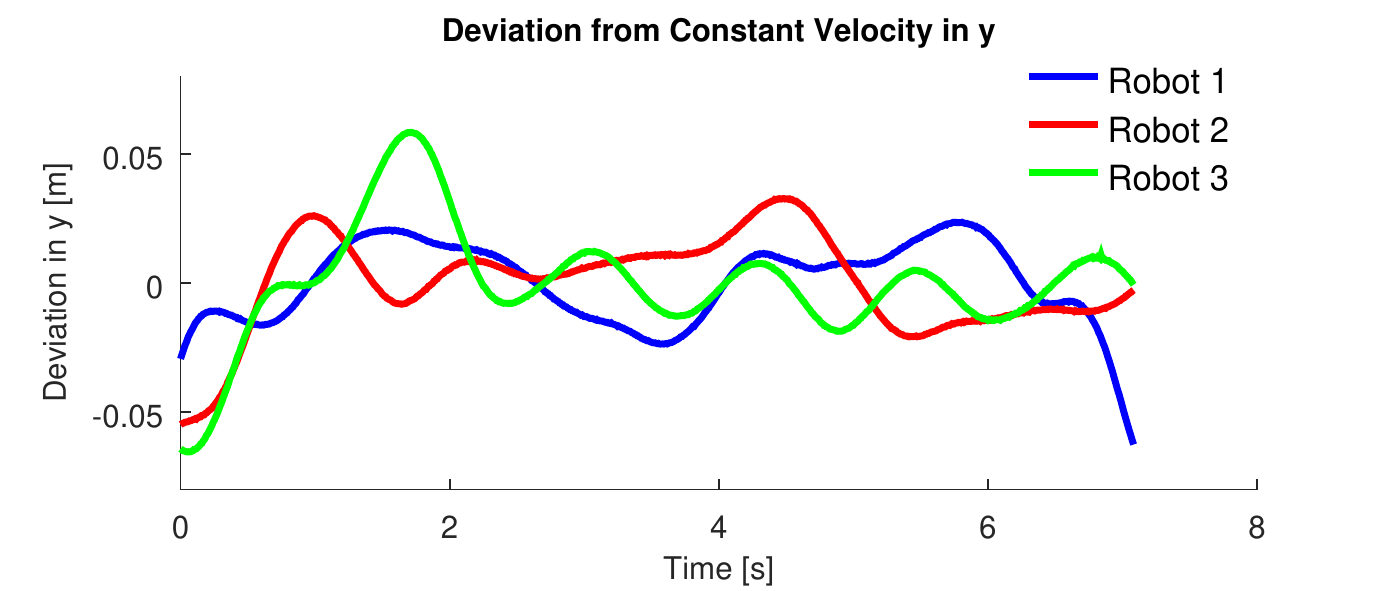}
\caption{}
\end{subfigure}
\caption{\textbf{Lennard-Jones} flight test: \textbf{(a), (b), (c)} See Fig.~\ref{fig:SD3} for description of these figures.}
\label{fig:LJ11}
\end{figure}

Post-experiment, we analyzed the robustness and stability of the collaborative transport algorithms, and quantified the cooperation between robots. For each flight test, we took seven seconds of linear translation data, which is depicted in Fig.~\ref{fig:SD3} and Fig.~\ref{fig:LJ11}. Fig.~\ref{fig:SD3}-a is the flight trajectory of one of the collaborative fabric transport tests using a spring-damper algorithm. Fig.~\ref{fig:LJ11}-a is the flight trajectory of one of the collaborative fabric transport tests using a Lennard-Jones potential. In a decentralized system, the behaviour of one robot can have great impacts on all of its neighbors. In particular, with non-deterministic forces arising from the flexible payload, it is expected that each robot must dynamically accomodate for its neighbours when they undergo instability. Thus, in a statistical analysis we should be able to measure these adjustments.

Parts (b) and (c) of figures \ref{fig:SD3} and \ref{fig:LJ11} illustrate the instabilities of each robot in a single dimension with respect to time. Without a shared payload the movement is expected to be directed only towards the goal, so these plots shows the deviation from that constant velocity trajectory. 
It can be observed that there is a similarity between the curves of Fig.~\ref{fig:SD3}, and of Fig.~\ref{fig:LJ11}. Synchronicity of UAV movements is ideal in this decentralized system, and it can be quantified by the average of the correlation coefficients of each pair of signals $\bar{\rho}$. This was done for each of the two dimensions $x$ and $y$, and for each flight test. The results are presented in Tab.~\ref{tab:cor}.

\begin{table}
\caption{Comparison of correlation coefficients for all flight tests.}
\begin{center}
\begin{tabular}{|c|c|c|c|} 
\hline
& $\bar{\rho_x}$ & $\bar{\rho_y}$ & $\frac{\bar{\rho_x} + \bar{\rho_y}}{2}$ \\
\hline
No Fabric & 0.19 & -0.17 & 0.01 \\
\hline
\multirow{5}{4em}{Spring-Damper}
& 0.74 & 0.44 & 0.59 \\
& 0.66 & 0.31 & 0.48 \\
& 0.60 & -0.02 & 0.29 \\
& 0.31 & 0.03 & 0.17 \\
& 0.29 & -0.06 & 0.12 \\
\hline
\multirow{5}{4em}{Lennard-Jones}
& 0.47 & 0.67 & 0.57 \\
& 0.43 & -0.08 & 0.17 \\
& 0.68 & 0.20 & 0.44 \\
& 0.55 & 0.28 & 0.41 \\
& 0.26 & -0.06 & 0.10 \\
\hline
\end{tabular}
\end{center}
\label{tab:cor}
\end{table}

For the Spring-Damper algorithm, the mean $\mu$ and the standard deviation $\sigma$ of the correlation coefficients from each flight test are
\begin{equation}
\mu_{SD} = 0.33,\ \ \sigma_{SD} = 0.20
\end{equation}
while for the Lennard-Jones algorithm,
\begin{equation}
\mu_{LJ} = 0.34,\ \ \sigma_{LJ} = 0.20
\end{equation}

The experiments show that the spring-damper and Lennard-Jones algorithms do in fact allow the robots to maintain safe, stable distances with similar efficacy. The non-linearity of the Lennard-Jones potential induces the same amount of cooperation between robots as the dampers that are present in the spring-damper control algorithm.
%
%


\section{FUTURE WORKS}
As mentioned at the beginning, this work is a first step along the way to get a safe fleet of micro-UAVs to dress a human. The decentralized approach was demonstrated to work well for such a scenario. Leveraging PyBuzz, other decentralized behaviours will be tested to succeed in creating autonomous flying dressing-aid. We showed our result to the designer and we will now try various shapes of clothing, designed specifically to be transport with micro-UAVs. Of course, the robustness of the algorithm will be enhanced, and together with infrared proximity detection, the interaction with users can begin to be studied.
\addtolength{\textheight}{-12cm}  


\section*{ACKNOWLEDGEMENT}

This project is based on an original idea of the designer and researcher Ying Gao, professor at the Fashion School of the University of Quebec in Montreal. 

\bibliographystyle{IEEEtran}
\bibliography{swarm,others}

\begin{thebibliography}{10}
\providecommand{\url}[1]{#1}
\csname url@samestyle\endcsname
\providecommand{\newblock}{\relax}
\providecommand{\bibinfo}[2]{#2}
\providecommand{\BIBentrySTDinterwordspacing}{\spaceskip=0pt\relax}
\providecommand{\BIBentryALTinterwordstretchfactor}{4}
\providecommand{\BIBentryALTinterwordspacing}{\spaceskip=\fontdimen2\font plus
\BIBentryALTinterwordstretchfactor\fontdimen3\font minus
  \fontdimen4\font\relax}
\providecommand{\BIBforeignlanguage}[2]{{%
\expandafter\ifx\csname l@#1\endcsname\relax
\typeout{** WARNING: IEEEtran.bst: No hyphenation pattern has been}%
\typeout{** loaded for the language `#1'. Using the pattern for}%
\typeout{** the default language instead.}%
\else
\language=\csname l@#1\endcsname
\fi
#2}}
\providecommand{\BIBdecl}{\relax}
\BIBdecl

\bibitem{Estevez2016}
J.~Estevez, J.~M. Lopez-Guede, and M.~Gra{\~{n}}a, ``{Particle Swarm
  Optimization Quadrotor Control for Cooperative Aerial Transportation of
  Deformable Linear Objects},'' \emph{Cybernetics and Systems}, vol.~47, no.
  1-2, pp. 4--16, 2016.

\bibitem{Ritz2012}
R.~Ritz, M.~W. M{\"{u}}ller, M.~Hehn, and R.~D'Andrea, ``{Cooperative
  quadrocopter ball throwing and catching},'' \emph{IEEE International
  Conference on Intelligent Robots and Systems}, pp. 4972--4978, 2012.

\bibitem{Ritz2013}
R.~Ritz and R.~D'Andrea, ``{Carrying a flexible payload with multiple flying
  vehicles},'' \emph{IEEE International Conference on Intelligent Robots and
  Systems}, pp. 3465--3471, 2013.

\bibitem{Dai2016}
Y.~Dai, Y.~Kim, S.~Wee, D.~Lee, and S.~Lee, ``{Symmetric caging formation for
  convex polygonal object transportation by multiple mobile robots based on
  fuzzy sliding mode control},'' \emph{ISA Transactions}, vol.~60, pp.
  321--332, 2016.

\bibitem{Loianno2018}
G.~Loianno and V.~Kumar, ``{Cooperative Transportation Using Small Quadrotors
  Using Monocular Vision and Inertial Sensing},'' \emph{IEEE Robotics and
  Automation Letters}, vol.~3, no.~2, pp. 680--687, 2018.

\bibitem{Preiss2017}
J.~A. Preiss, W.~H{\"{o}}nig, G.~S. Sukhatme, and N.~Ayanian, ``{Crazyswarm: A
  large nano-quadcopter swarm},'' \emph{IEEE International Conference on
  Robotics and Automation ({\{}ICRA{\}})}, pp. 3299--3304, 2017.

\bibitem{Furci2015}
M.~Furci, G.~Casadei, R.~Naldi, R.~G. Sanfelice, and L.~Marconi, ``{An
  open-source architecture for control and coordination of a swarm of
  micro-quadrotors},'' \emph{2015 International Conference on Unmanned Aircraft
  Systems, ICUAS 2015}, pp. 139--146, 2015.

\bibitem{Honig2017}
W.~H{\"{o}}nig and N.~Ayanian, ``{Flying multiple UAVs using ROS},''
  \emph{Studies in Computational Intelligence}, vol. 707, pp. 83--118, 2017.

\bibitem{Pinciroli2016b}
C.~Pinciroli and G.~Beltrame, ``{Swarm-Oriented Programming of Distributed
  Robot Networks},'' \emph{Computer}, vol.~49, no.~12, pp. 32--41, 2016.

\bibitem{Stoy2001}
K.~St{\o}y, ``{Using situated communication in distributed autonomous mobile
  robots},'' \emph{Proceedings of the 7th Scandinavian Conference on Artificial
  Intelligence}, pp. 44--52, 2001.

\bibitem{Behnel2011}
S.~Behnel, R.~Bradshaw, C.~Citro, L.~Dalcin, D.~S. Seljebotn, and K.~Smith,
  ``{Cython: The best of both worlds},'' \emph{Computing in Science and
  Engineering}, vol.~13, no.~2, pp. 31--39, 2011.

\bibitem{Estevez2015}
J.~Est{\'{e}}vez, J.~M. Lopez-Guede, and M.~Gra{\~{n}}a, ``{Quasi-stationary
  state transportation of a hose with quadrotors},'' \emph{Robotics and
  Autonomous Systems}, vol.~63, no.~P2, pp. 187--194, 2015.

\bibitem{Belkacem2016}
K.~Belkacem and F.~Cherif, ``{Swarm Robots Circle Formation via a Virtual
  Viscoelastic Control Model},'' in \emph{International Conference on
  Modelling, Identification and Control}, ALgiers, 2016, pp. 725--730.

\bibitem{Rimon1992}
E.~Rimon and D.~Koditschek, ``{Exact Robot Navigation Using Artificial
  Potential Functions},'' \emph{Robotics and Automation, IEEE}, vol.~8, no.~5,
  pp. 501--518, 1992.

\bibitem{Reif1999}
J.~H. Reif and H.~Wang, ``{Robotics and Autonomous Systems Social potential
  fields: A distributed behavioral control for autonomous robots*},''
  \emph{Robotics and Autonomous Systems}, vol.~27, pp. 171--194, 1999.

\end{thebibliography}

\end{document}